\documentclass[conference]{IEEEtran}
\IEEEoverridecommandlockouts
\usepackage{cite}
\usepackage[numbers,sort&compress,square]{natbib}
\usepackage{url}
\usepackage{graphicx}
\usepackage{textcomp}
\usepackage{xcolor}
\usepackage[most]{tcolorbox}

\usepackage{amsmath,amssymb,amsfonts}
\usepackage[linesnumbered,ruled,noend]{algorithm2e}
\usepackage{array}
\usepackage{multicol}
\usepackage{multirow}
\usepackage{url}
\usepackage{hyperref}

\makeatletter

\newcommand{\removelatexerror}{\let\@latex@error\@gobble}
\makeatother

\newcolumntype{M}[1]{>{\centering\arraybackslash}m{#1}}

\SetKwInput{KwInput}{Input}
\SetKwInput{KwOutput}{Output}

\def\BibTeX{{\rm B\kern-.05em{\sc i\kern-.025em b}\kern-.08em
    T\kern-.1667em\lower.7ex\hbox{E}\kern-.125emX}}
\begin{document}

\title{Critical Sentence Identification in Legal Cases Using Multi-Class Classification \\
}
\author{
\IEEEauthorblockN{
Sahan Jayasinghe,
Lakith Rambukkanage, Ashan Silva, Nisansa de Silva, Amal Shehan Perera
}
\IEEEauthorblockA{
\textit{Department of Computer Science \& Engineering} \\
\textit{University of Moratuwa}
}
\IEEEauthorblockA{Email : sahanjayasinghe.17@cse.mrt.ac.lk}

}

\date{Nov 2021}

\maketitle

\begin{abstract}
Inherently, the legal domain contains a vast amount of data in text format. Therefore it requires the application of Natural Language Processing (NLP) to cater to the analytically demanding needs of the domain. The advancement of NLP is spreading through various domains, such as the legal domain, in forms of practical applications and academic research. Identifying critical sentences, facts and arguments in a legal case is a tedious task for legal professionals. In this research we explore the usage of sentence embeddings for multi-class classification to identify critical sentences in a legal case, in the perspective of the main parties present in the case. In addition, a task-specific loss function is defined in order to improve the accuracy restricted by the straightforward use of categorical cross entropy loss.
\end{abstract}


\section{Introduction}
\label{section :intro}

Text data is a form of unstructured data which inherently makes it harder to analyse compared to structured or numerical data. Moreover when the quantity of text data increases, the difficulty increases further. With the advancement in Natural Language Processing (NLP), many researches have been carried out in various domains, as academic research and for practical applications to tackle these problems. The legal domain is such a domain in which research interests have come up over the recent past, and many research have been carried out considering its different aspects. Party based sentiment analysis~\cite{rajapaksha2020rule,mudalige2020sigmalaw,rajapaksha2021sigmalaw}, extracting parties of a legal case~\cite{samarawickrama2020party,de2020legal,samarawickrama2021identifying}, detecting critical sentences of a court case and predicting the outcome of a court case are such important aspects. 

\subsection{Importance of NLP in the Legal Domain}
Natural Language Processing (NLP) fundamentally provides with different techniques to extract information from textual data. Not only, NLP can be used to process large amount of text data to extract information, but also, it performs well in large amounts of textual data. Legal domain is such a domain where textual data is piling up on a daily basis and repetitive analytical work is done on these textual data by legal professionals, regularly. Therefore it is important to explore the possibilities of leveraging NLP techniques to reduce the amount of manual work in the legal domain.

\subsection{Embeddings in NLP}
In NLP, word embeddings are a key concept where the semantic meaning of a word is captured in the form of a vector of a particular dimension. The same notion have been adapted to extract the semantic meaning of a sentence, through sentence embeddings. In the legal domain a word alone can mean many things but only a sentence would make sufficient semantic meaning.

\subsection{Case Law}
In the legal domain, Case Law is the usage of past court cases to support or counter arguments of an ongoing court case rather than using constitutions, statutes and other formal legal definitions~\cite{CaseLaw}. Legal professionals make use of decisions given in the past in similar court cases to support their claims. For this purpose they go through the case law documents to find similar cases and valuable arguments in them. 

\subsection{Legal Parties in a Court Case}
A legal party in a court case can be a person, a group of people of an entity. In any given court case, two main legal parties can be identified~\cite{LegalParty}. The party filing the case to the court is know as either \textit{petitioner} or \textit{plaintiff} and in criminal cases as \textit{prosecutor}. The party responding of defending themselves are called  \textit{defendant}, \textit{respondent} or \textit{accused}. Note that there maybe \textit{third party}  entities or persons that do not belong to either of the main parties. From here onward the two main parties will be referred to as \textit{petitioner} and \textit{defendant} for ease of comprehension. Almost all the sentences, facts and arguments in a court case are presented to support or oppose the two main parties.

\subsection{Critical Sentences in Court Case}
Among many other tedious tasks, legal professionals spend significant amount of time reading case law documents, trying to extract important facts to support or counter an ongoing case. These documents are relatively complex in language, and the amount of documents they have to go through expand each year as well. If this repetitive tedious task of extracting important facts or arguments could be done through NLP it would reduce the manual work and save a lot of time.

The main focus of this research is to identify or categorize the critical facts and arguments in terms of how they support or oppose the main parties of a court case:petitioner and defendant in the court with the support of the known decisions of case law documents. The sentence embedding technique, the training dataset used for this research and other related work is discussed in Section \ref{section :related_work}. Section \ref{section: methodology} consists of the methodology used to formulate the model. The experiments and results are elaborated in Section \ref{section: results} and conclusions and future work are discussed in Section \ref{section: conclusion}.

\section{Related Work}
\label{section :related_work}

\subsection{Word Embeddings}

Bidirectional Encoder Representations from Transformers (BERT)~\cite{devlin2018bert} is a word embedding technique, designed with the intention of removing the concerns on using unidirectional language models. This solves the issue with the restriction on choice of architecture in using pre-trained models with fine-tuning or feature based applications. The BERT model architecture is a Multi layer Bidirectional Transformer and uses two pre-training tasks: Masked Language Model (MLM) and next sentence prediction (NSP). Fine tuning is mentioned as relatively straightforward but computationally expensive. The model has achieved higher scores in General Language Understanding Evaluation (GLUE) benchmark~\cite{wang2018glue} compared to many other models available at the time.


RoBERTa~\cite{liu2019roberta} is an advanced model trained on more data compared to BERT~\cite{devlin2018bert}. BERT is trained on 16GB of text data whereas RoBERTa uses 160GB text data.  The model is evaluated using the GLUE benchmark~\cite{wang2018glue}. BERT uses static masking where the words are randomly chosen and masked once, but RoBERTa uses dynamic masking which has improved the accuracy marginally on GLUE evaluation. They have considered the Next sentence prediction (NSP) used in BERT and evaluated with and without NSP loss under four criteria.
With the evaluation against GLUE benchmarks, they have seen NSP loss inclusion is lowering accuracy. Out of the three experimented batch sizes 256, 2000, 8000, they have seen 2000 performs best. Since the method used for encoding have not impacted much on the accuracy, they have used Byte-Pair Encoding.

\subsection{Sentence Embedding}


In the Sentence-BERT model, researchers~\citet{reimers2019sentence} have used the Stanford Natural Language Inference (SNLI) dataset for training the model, which is labeled as contradiction, entailment and neutral. They have used pooling techniques on both RoBERTa~\cite{liu2019roberta} and BERT~\cite{devlin2018bert} outputs to get uniform length sentence embeddings. Three objective functions have been used for optimization:  Classification objective function which uses the difference of vectors passed through a SoftMax function, Regression objective function where cosine similarity is used for loss calculation, and Triplet objective function which considers two positive and negative sentences with respect to an anchor sentence and minimize the distant from anchor to positive sentence. They have finally concluded to use the SoftMax-classification objective function, with Adam optimizer. They have outperformed many SemEval tasks compared to InferSent~\cite{conneau2017supervisedInferSent} and Universal Sentence Encoder~\cite{cer2018universal}.

InferSent~\cite{conneau2017supervisedInferSent} is a sentence embedding model based on supervised learning approach with bidirectional LSTMs and Max pooling. According to the paper this is the first research that outperforms the models based on unsupervised embedding techniques that were available at that time. They have used multiple encoding methods to encode sentences to fixed lengths: LSTMs and GRUs, Bidirectional LSTMs with Max/mean pooling, Self-attentive networks, Hierarchical convolution. Bidirectional LSTMs with max pooling have performed the best. Also, the results have increased as the embedding size increases. 



Universal Sentence Encoder~\cite{cer2018universal} is a sentence encoding model based on a research conducted by researchers at google, which makes use of transfer learning. They have used two models for this, one of them which uses Transformer Architecture and other depends on a Deep Averaging Network. They have identified that these embeddings can be fine tuned for specific tasks easily. The transformer-based encoder has achieved better accuracy and has more general-purpose embedding.


In the research carried out by~\citet{tang2014learningTwitter} addresses the requirement for the representation of sentiment in embedding without considering semantics only which is identified by them as Sentiment specific word embedding. They have developed three Neural Networks to get sentiment specific word embeddings. Researchers have trained it for 10 million tweets of which emoticons are used as the label for the data in the form of distant supervision. They have first tried out two models with strict polarity values of 1,0 and then have loosen restrictions on predictions to decimals. Finally they have created a unified model which outputs the sentiment and semantics both as embeddings. 
They have defined an evaluation to the two outputs separately (semantics and sentiment) and then defined a combined loss function. 

\subsection{Party Based Sentiment Analysis Dataset}

As described in the work of \citet{mudalige2020sigmalaw} NLP in the legal domain lacks proper datasets for party based sentiment analysis. This issue is addressed by annotating party based sentiment to a dataset originally extracted from the United States Supreme Court consisting of nearly 2000 sentences. The sentiment annotation is done with three notions in mind: Two main parties of a case which are petitioner and defendant, Sentiment for each party within the sentence labeled as positive (+1), neutral (0) and negative (-1), overall sentiment of the sentence regardless of the parties labeled as positive (+1), neutral (0) and negative (-1).

\subsection{Cross Entropy Loss}
As explained in the work of~\citet{zhang2018generalized}, classification using Mean Absolute Error (MAE) as the loss function works only under few assumptions. When the classification labels of the dataset becomes noisy MAE starts to work poorly, where as Cross Entropy Loss or tuning of Cross entropy loss works well with noisy data. \citet{zhang2018generalized} have experimented modified Cross Entropy loss functions with synthetically added noise for 3 datasets at different noise levels and all of them have outperformed MAE.

\subsection{Critical Sentence identification}


In the work of \citet{glaser2018classifying} the classification of sentences in legal contracts based on 9 different semantic classes are explored. The research experiment has been conducted for legal texts in German language. First the Machine Learning (ML) models have been trained for sentences extracted from the German Civil code and later used on legal contracts to extract the corresponding critical sentences in a contract.

Researchers \citet{jagadeesh2005sentence} have proposed a sequential process to summarize documents and then extract important sentences. For this purpose, in the first part they have employed syntactic parsing with Named Entity Recognition (NER) and Parts of Speech (PoS) Tagging followed by feature extraction based on the NER, PoS tags, word frequencies etc. Afterwards sentences are scored based on the features and ranked accordingly. The system allows to extract sentences based on a query by generating a coherence score is a key element.

\citet{hirao2002extracting} have explored the sentence extraction aspect by using Support Vector Machines (SVM). The approach to extracting the critical sentences from a document is similar to text summarizing, and classifying the sentences based on two classes of important and unimportant. They have been able to beat three other approaches for text summarizing at the time as well.

\section{Methodology}
\label{section: methodology}
Initially exploratory analysis was carried out on sentences with only the decision annotated and it was identified that the dataset needs to be more detailed for the model to learn. With this insight a proper dataset was needed to be prepared for the task. The dataset for this research is an extension of the Party Based Sentiment Analysis (PBSA)  dataset created by \citet{mudalige2020sigmalaw} which is publicly available at OSF\footnote{SigmaLaw ABSA Dataset: \url{https://osf.io/37gkh/}}. The PBSA dataset consists of 1822 exact sentences and meaningful sub sentences extracted from 25 case law documents of United States Supreme Court by \citet{sugathadasa2017synergistic}, along with the sentiment for each party present in them and the overall sentiment.

\subsection{Dataset Preparation}
\label{subsection_data_anon}
We used 1822 sentences of the PBSA dataset developed by \citet{mudalige2020sigmalaw} along with the party and sentiment annotations. The dataset was extended by adding the decision of each court case for each sentence. We identified four classes that a sentence can be belonged to, based on the winning party of the court case and the impact of the sentence towards the party. A sentence can be classified as follows.

\begin{enumerate}
    \item Petitioner lose and has negative impact towards petitioner (Petitioner lose and negative)
    \item Petitioner lose and has positive impact towards petitioner (Petitioner lose and positive)
    \item Petitioner win and has negative impact towards petitioner (Petitioner win and negative)
    \item Petitioner win and has positive impact towards petitioner (Petitioner win and positive)
\end{enumerate}

For simplicity, we will refer the above categories in a shorter form as mentioned in the brackets.
According to above classes, a sentence can be considered as critical when it has,
\begin{itemize}
    \item A positive impact towards petitioner in a case document where petitioner won
    \item A negative impact toward petitioner in a case document where petitioner lost
\end{itemize}

In PBSA dataset, the parties mentioned in each sentence are annotated with their sentiment labels. We used those sentiment labels to calculate the impact towards petitioner for each sentence.

\begin{figure}[!htb]
\begin{tcolorbox}[colbacktitle=black, title=Case Sentence 1 from Jae Lee V. US \cite{1977lee}]
\label{textbox: case_sentence_1}
\small{After obtaining a warrant, the officials searched Lee's house, where they found drugs, cash, and a loaded rifle.}
\end{tcolorbox}
\end{figure}

In the case sentence 1, the petitioner is Lee and the defendant party is represented by the officials. PBSA dataset contains the sentiment labels for the words \emph{Lee}, \emph{officials} and \emph{they} as follows.
\begin{itemize}
    \item Petitioners: Lee – Negative (-1)
    \item Defendants: Officials – Positive (+1), they -  Positive (+1)
\end{itemize}

According to the sentiment labels, the impact towards petitioner in the sentence \ref{textbox: case_sentence_1} can be annotated as negative.

Whenever there is a sentence which mentions only the members of the defendant party, we consider the inverse sentiment of the defendants as the impact towards petitioner for the purpose of categorizing the sentence according to the 4 classes.
Within the 1822 sentences, there were 214 sentences which reflected a neutral sentiment towards any of the parties. We removed those sentences since those are insignificant for the sentence importance prediction and also they would lead to a high class imbalance due to the abundance. In case of providing a sentence with a neutral sentiment to the model trained on the following classes, the output probabilities should not significantly be bias to any class.

Remaining 1608 sentences are labeled as Petitioner win/lose regarding the decision of the case document the respective sentence belongs to.
Sentence counts of each class are:
\begin{itemize}
    \item Petitioner lose and negative – 226
    \item Petitioner lose and positive – 230
    \item Petitioner win and negative – 687
    \item Petitioner win and positive – 465
\end{itemize}

Since most of the US Supreme court cases are appeals by petitioners, there is a marginally higher number of sentences which affect negatively towards petitioner.




\subsection{Multi-class Classification Model}
The proposed architecture for the multi-class classification task consists of 3 components: Token Embedding model, a Mean Pooling Layer and a Dense Layer. For the tokenization process, we used pretrained BertTokenizer containing vocabulary of 30000 words and sub-words. For the Embedding model, we used Bert-base-cased model \footnote{bert-base-cased: \url{https://huggingface.co/bert-base-cased}} pretrained on Wikipedia text. Final Dense layer consists of nodes equal to number of classes.

\begin{figure}[!htb]
    \centering
    \includegraphics[width=0.5\textwidth]{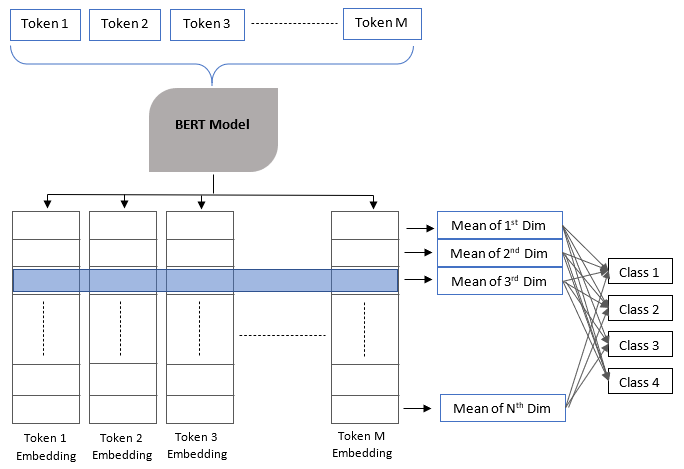}
    \caption{Model Architecture}
    \label{fig:model_diagram}
\end{figure}

Bert Tokenizer is configured to generate tokens for the words and sub-words identified in the given input text and create a padded token sequence of 128 long. An input mask is also generated so that the Bert model can distinguish the tokens representing the text. Bert model generates a 768-dimensional embedding for a single token. To obtain a unified representation for the input text sequence, there are 3 techniques mentioned by \citet{reimers2019sentence}: using the [CLS] token embedding, taking the mean of token embeddings at each dimension (Mean Pooling), taking maximum of the token embeddings at each dimension (Max Pooling). In this research, we used Mean Pooling to obtain a generalized representation for the sentences provided to the model as input.

According to \citet{devlin2018bert}, Fine-tuning BERT for downstream classification tasks would often yield better results rather than using a pretrained word embeddings for a particular domain.
Model is implemented using Sentence Transformers Git repository\footnote{Sentence Transformers: \url{https://github.com/UKPLab/sentence-transformers}} and the codes for classification task is implemented using PyTorch.

\subsection{Task-specific Loss Function}
Categorical Cross Entropy loss is the most promising loss function for multi-class classification tasks. Cross Entropy loss is calculated using only the probability of the labeled class. But we defined a new loss function which takes into account the probabilities of rest of the classes instead of the labeled class.

The classification task discussed in this paper contains four classes that can be separated into two polarities depending on the decision of the court case.

\begin{table}[h!]
\begin{center}
\caption{Class Polarity}
\begin{tabular}{ c|c }
    Petitioner lose and negative & Petitioner win and negative \\ 
    Petitioner lose and positive & Petitioner win and positive \\
\end{tabular}
\end{center}
\label{table: class_polarity}
\end{table}

We defined the loss function as in Algorithm \ref{algo:loss_fn}, which takes into account the polarity of classes.

\begin{figure}[!ht]
    \removelatexerror
    \begin{algorithm}[H]
    \DontPrintSemicolon
    \SetKwFunction{FOppositeClassWeightedLoss}{OppositeClassWeightedLoss}
    \KwInput{Output of the model, Class Label, Weight that multiplies the losses of opposite classes}
    \KwOutput{loss value}
    
    \SetKwProg{Fn}{Function}{:}{}
    \Fn{\FOppositeClassWeightedLoss{$ModelOutput$, $Label$, $OppositeClassWeight$}}
    {
        NumClasses := Length(ModelOutput)\;
        MidClassIndex := (NumClasses + 1) div 2\;
        
        SoftmaxOutput := \emph{Softmax}(ModelOutput)\;
        Loss := 0\;
        
        \For{ClassIndex:= 0 : NumClasses-1}
        {
            \If{Label = ClassIndex}
            {
                Weight := 0\;
            }
            \ElseIf{Label $<$ MidClassIndex}
            {
                \If{ClassIndex $\geq$ MidClassIndex}{Weight := OppositeClassWeight}
                \Else{Weight := 1}
            }
            \Else
            {
                \If{ClassIndex $<$ MidClassIndex}{Weight := OppositeClassWeight}
                \Else{Weight := 1}
            }
            Loss := Loss + Weight * \emph{log}(1 - SoftmaxOutput[ClassIndex])
        }
         \KwRet \emph{Loss}
    }
    \textbf{end Function}
    \caption{Loss Function for case sentence classification}
    \label{algo:loss_fn}
    \end{algorithm}
\end{figure}

The goal of the loss function is to penalize more on the probabilities of opposite classes. For an example, when the label is Petitioner lose and negative, the opposite classes are Petitioner win and negative, Petitioner win and positive. In the context of legal domain, predicting a sentence to be in a case document where petitioner lost, while that sentence exists in a case where petitioner won is a severe error when compared to predicting the wrong sentiment (impact).

Algorithm \ref{algo:loss_fn} takes in the output of each node at the last layer of the model, then Softmax, shown in equation~\ref{eq:softmax}, is applied to obtain probabilities of each class, where $x_i$ is the output of the $i$th node.  

\begin{equation}
    Softmax(x_i) = \frac{e^{x_i}}{\sum_{j}e^{x_j}}
    \label{eq:softmax}
\end{equation}

\textit{MidClassIndex} field represents the class index which switches the polarity. According to above four classes, class indices 0 and 1 represents the petitioner lose polarity, class indices 2 and 3 represents the petitioner win polarity. Depending on the label of the case sentence, a weight is dynamically applied to the probability loss of the opposite classes. Weight for the labeled class is 0. Weight for the classes which has the same polarity as the labeled class is 1. 

The weight applied for opposite classes is configured as a hyper-parameter which we have experimented in Section \ref{section: results}. When the predicted probabilities of the opposite classes are high, the loss becomes a higher value as shown in Equation~\ref{eq:loss}, where $C_i$ is the $i$th class, $L_{c_i}$ is the loss calculated for the $i$th class, $W_i$ is the current weight associated with the $i$th class, and $P_{c_i}$ is the predicted probability of the $i$th class.

\begin{equation}
   L_{c_i} = W_{c_i} * log(1 - P_{c_i})
    \label{eq:loss}
\end{equation}

Total loss for an input sentence is the accumulative loss of each class \textit{i}.

\section{Experiments and Results}
\label{section: results}
From the 1608 labeled case sentences we prepared as mentioned in Section \ref{subsection_data_anon}, Train, Validation and Test set splits are created according to Table \ref{table: dataset_stat}. Over-sampling and under-sampling techniques are used to mitigate the class imbalance in the Train set.

\begin{table}[h!]
\begin{center}
\caption{Dataset Statistics}
\begin{tabular}{ |M{3cm}||M{1cm}|M{1cm}|M{0.7cm}|M{0.7cm}| }
    \hline
    Class & Original Train & Over-sampled Train & Valida-tion & Test \\
    \hline
    \hline
    Petitioner lose \& negative & 176 & 547 & 25 & 25 \\
    \hline
    Petitioner lose \& positive & 180 & 547 & 25 & 25 \\
    \hline
    Petitioner win \& negative & 547 & 547 & 70 & 70 \\
    \hline
    Petitioner win \& positive & 365 & 547 & 50 & 50 \\
    \hline
\end{tabular}
\end{center}
\label{table: dataset_stat}
\end{table}

In the over-sampling approach, samples from the 3 classes with lower sentence counts are duplicated to match the count of the \textit{Petitioner win \& negative} class of the train set. In contrast, the under-sampling method reduces the examples of 3 classes which consist of higher counts, to match with the class with lowest number of examples. These 2 combinations of the dataset are used to train the classification model, first with categorical cross entropy loss, then with the task-specific loss function. We have experimented the opposite class loss weight hyper-parameter and the metrics comparison is displayed in Table \ref{table: metrics}.

\begin{table}[h!]
\centering
\caption{Performance Metrics}
\begin{tabular}{|M{1cm}|M{1cm}||M{1cm}|M{1cm}||M{1cm}|M{1cm}|}
    \hline
    \multirow{2}{1cm}{\textbf{Loss Function}} & \multirow{2}{1cm}{\textbf{Opposite Class Loss weight}} & \multicolumn{2}{c||}{Over-sampled} & \multicolumn{2}{c|}{Under-sampled} \\
    \cline{3-6}
    & & \textbf{Accuracy (\%)} & \textbf{Macro-F1 (\%)} & \textbf{Accuracy (\%)} & \textbf{Macro-F1 (\%)} \\
    \hline
    Categori-cal Cross Entropy & N/A & 68.82 & 67.86 & 58.24 & 59.99 \\
    \hline
    \multirow{8}{1cm}{Task-Specific Loss Function} & 1 & 72.94 & 70.39 & 63.53 & 64.68 \\
    \cline{2-6}
    & 2 & 73.53 & 72.02 & 61.18 & 60.49 \\
    \cline{2-6}
     & 3 & 71.76 & 70.97 & 62.94 & 63.68 \\
    \cline{2-6}
     & 4 & 74.12 & \textbf{73.62} & 63.53 & 64.83 \\
    \cline{2-6}
     & 5 & 74.12 & 73.26 & \textbf{65.88} & \textbf{65.66} \\
    \cline{2-6}
    & 6 & \textbf{75.29} & 73.24 & 60.00 & 60.52 \\
    \cline{2-6}
    & 7 & 74.71 & 73.57 & 59.41 & 60.90 \\
    \cline{2-6}
    & 8 & 71.18 & 70.67 & 59.41 & 61.11 \\
    \hline
\end{tabular}
\label{table: metrics}
\end{table}

Each model is trained for maximum 8 epochs to avoid over-fitting. Afterwards the model weights with the best validation accuracy is used to evaluate on the test set. According to Table \ref{table: metrics}, \textit{Task-Specific loss function} provided better optimization for the sentence classification when compared to the categorical cross entropy loss. \textit{Opposite class loss weight} is configured as a hyper-parameter for training and the values 4, 5, 6 showed the best results for the test set. The increased accuracy of the task specific loss function can be intuitively explained as a result of considering domain dependent case decision polarity. 

\section{Conclusion and Future Work}
\label{section: conclusion}

Thus far, in this study we have explored the classification of critical sentences considering the decision of the court case and the impact on the petitioner party. With the use of transformer based embeddings for the input sentences and the task-specific loss function, a better classification results have been obtained. Task-specific loss function outperformed the state-of-the-art categorical cross entropy loss, when training the classification model in the context of legal domain. Automating the identification of critical sentences with the task specific loss function and more annotated data to improve model performance, would hopefully reduce the manual and analytical demanding work of the legal professionals. As future work, we expect to develop a Natural Language Inference (NLI) dataset along with semantic similarity scores (STS) for the legal domain, which will then be used to fine-tune and evaluate a sentence embeddings model.

\bibliographystyle{IEEEtranN}
\bibliography{SigmaLaw,ICIIS2021-citations}

\begin{thebibliography}{21}
\providecommand{\natexlab}[1]{#1}
\providecommand{\url}[1]{#1}
\csname url@samestyle\endcsname
\providecommand{\newblock}{\relax}
\providecommand{\bibinfo}[2]{#2}
\providecommand{\BIBentrySTDinterwordspacing}{\spaceskip=0pt\relax}
\providecommand{\BIBentryALTinterwordstretchfactor}{4}
\providecommand{\BIBentryALTinterwordspacing}{\spaceskip=\fontdimen2\font plus
\BIBentryALTinterwordstretchfactor\fontdimen3\font minus
  \fontdimen4\font\relax}
\providecommand{\BIBforeignlanguage}[2]{{%
\expandafter\ifx\csname l@#1\endcsname\relax
\typeout{** WARNING: IEEEtranN.bst: No hyphenation pattern has been}%
\typeout{** loaded for the language `#1'. Using the pattern for}%
\typeout{** the default language instead.}%
\else
\language=\csname l@#1\endcsname
\fi
#2}}
\providecommand{\BIBdecl}{\relax}
\BIBdecl

\bibitem[Rajapaksha et~al.(2020)Rajapaksha, Mudalige, Karunarathna, de~Silva,
  Rathnayaka, and Perera]{rajapaksha2020rule}
I.~Rajapaksha, C.~R. Mudalige, D.~Karunarathna, N.~de~Silva, G.~Rathnayaka, and
  A.~S. Perera, ``Rule-based approach for party-based sentiment analysis in
  legal opinion texts,'' \emph{arXiv preprint arXiv:2011.05675}, 2020.

\bibitem[Mudalige et~al.(2020)Mudalige, Karunarathna, Rajapaksha, de~Silva,
  Ratnayaka, Perera, and Pathirana]{mudalige2020sigmalaw}
C.~R. Mudalige, D.~Karunarathna, I.~Rajapaksha, N.~de~Silva, G.~Ratnayaka,
  A.~S. Perera, and R.~Pathirana, ``Sigmalaw-absa: Dataset for aspect-based
  sentiment analysis in legal opinion texts,'' \emph{arXiv preprint
  arXiv:2011.06326}, 2020.

\bibitem[Rajapaksha et~al.(2021)Rajapaksha, Mudalige, Karunarathna, Silva,
  Perera, and Ratnayaka]{rajapaksha2021sigmalaw}
I.~Rajapaksha, C.~R. Mudalige, D.~Karunarathna, N.~d. Silva, A.~S. Perera, and
  G.~Ratnayaka, ``{Sigmalaw PBSA-A Deep Learning Model for Aspect-Based
  Sentiment Analysis for the Legal Domain},'' in \emph{International Conference
  on Database and Expert Systems Applications}.\hskip 1em plus 0.5em minus
  0.4em\relax Springer, 2021, pp. 125--137.

\bibitem[Samarawickrama et~al.(2020)Samarawickrama, de~Almeida, de~Silva,
  Ratnayaka, and Perera]{samarawickrama2020party}
C.~Samarawickrama, M.~de~Almeida, N.~de~Silva, G.~Ratnayaka, and A.~S. Perera,
  ``{Party Identification of Legal Documents using Co-reference Resolution and
  Named Entity Recognition},'' in \emph{2020 IEEE 15th International Conference
  on Industrial and Information Systems (ICIIS)}.\hskip 1em plus 0.5em minus
  0.4em\relax IEEE, 2020, pp. 494--499.

\bibitem[de~Almeida et~al.(2020)de~Almeida, Samarawickrama, de~Silva,
  Ratnayaka, and Perera]{de2020legal}
M.~de~Almeida, C.~Samarawickrama, N.~de~Silva, G.~Ratnayaka, and A.~S. Perera,
  ``{Legal Party Extraction from Legal Opinion Text with Sequence to Sequence
  Learning},'' in \emph{2020 20th International Conference on Advances in ICT
  for Emerging Regions (ICTer)}.\hskip 1em plus 0.5em minus 0.4em\relax IEEE,
  2020, pp. 143--148.

\bibitem[Samarawickrama et~al.(2021)Samarawickrama, de~Almeida, Perera,
  de~Silva, and Ratnayaka]{samarawickrama2021identifying}
C.~Samarawickrama, M.~de~Almeida, A.~S. Perera, N.~de~Silva, and G.~Ratnayaka,
  ``Identifying legal party members from legal opinion texts using natural
  language processing,'' EasyChair, Tech. Rep., 2021.

\bibitem[Cas()]{CaseLaw}
``Case law,'' \url{https://www.law.cornell.edu/wex/case_law}, accessed:
  2021-05-27.

\bibitem[Leg()]{LegalParty}
``Legal party,'' \url{https://www.law.cornell.edu/wex/party}, accessed:
  2021-05-27.

\bibitem[Devlin et~al.(2018)Devlin, Chang, Lee, and Toutanova]{devlin2018bert}
J.~Devlin, M.-W. Chang, K.~Lee, and K.~Toutanova, ``Bert: Pre-training of deep
  bidirectional transformers for language understanding,'' \emph{arXiv preprint
  arXiv:1810.04805}, 2018.

\bibitem[Wang et~al.(2018)Wang, Singh, Michael, Hill, Levy, and
  Bowman]{wang2018glue}
A.~Wang, A.~Singh, J.~Michael, F.~Hill, O.~Levy, and S.~R. Bowman, ``Glue: A
  multi-task benchmark and analysis platform for natural language
  understanding,'' \emph{arXiv preprint arXiv:1804.07461}, 2018.

\bibitem[Liu et~al.(2019)Liu, Ott, Goyal, Du, Joshi, Chen, Levy, Lewis,
  Zettlemoyer, and Stoyanov]{liu2019roberta}
Y.~Liu, M.~Ott, N.~Goyal, J.~Du, M.~Joshi, D.~Chen, O.~Levy, M.~Lewis,
  L.~Zettlemoyer, and V.~Stoyanov, ``Roberta: A robustly optimized bert
  pretraining approach,'' \emph{arXiv preprint arXiv:1907.11692}, 2019.

\bibitem[Reimers and Gurevych(2019)]{reimers2019sentence}
N.~Reimers and I.~Gurevych, ``Sentence-bert: Sentence embeddings using siamese
  bert-networks,'' \emph{arXiv preprint arXiv:1908.10084}, 2019.

\bibitem[Conneau et~al.(2017)Conneau, Kiela, Schwenk, Barrault, and
  Bordes]{conneau2017supervisedInferSent}
A.~Conneau, D.~Kiela, H.~Schwenk, L.~Barrault, and A.~Bordes, ``Supervised
  learning of universal sentence representations from natural language
  inference data,'' \emph{arXiv preprint arXiv:1705.02364}, 2017.

\bibitem[Cer et~al.(2018)Cer, Yang, Kong, Hua, Limtiaco, John, Constant,
  Guajardo-C{\'e}spedes, Yuan, Tar, et~al.]{cer2018universal}
D.~Cer, Y.~Yang, S.-y. Kong, N.~Hua, N.~Limtiaco, R.~S. John, N.~Constant,
  M.~Guajardo-C{\'e}spedes, S.~Yuan, C.~Tar \emph{et~al.}, ``Universal sentence
  encoder,'' \emph{arXiv preprint arXiv:1803.11175}, 2018.

\bibitem[Tang et~al.(2014)Tang, Wei, Yang, Zhou, Liu, and
  Qin]{tang2014learningTwitter}
D.~Tang, F.~Wei, N.~Yang, M.~Zhou, T.~Liu, and B.~Qin, ``Learning
  sentiment-specific word embedding for twitter sentiment classification,'' in
  \emph{Proceedings of the 52nd Annual Meeting of the Association for
  Computational Linguistics (Volume 1: Long Papers)}, 2014, pp. 1555--1565.

\bibitem[Zhang and Sabuncu(2018)]{zhang2018generalized}
Z.~Zhang and M.~R. Sabuncu, ``Generalized cross entropy loss for training deep
  neural networks with noisy labels,'' in \emph{32nd Conference on Neural
  Information Processing Systems (NeurIPS)}, 2018.

\bibitem[Glaser et~al.(2018)Glaser, Scepankova, and
  Matthes]{glaser2018classifying}
I.~Glaser, E.~Scepankova, and F.~Matthes, ``Classifying semantic types of legal
  sentences: Portability of machine learning models,'' in \emph{Legal Knowledge
  and Information Systems}.\hskip 1em plus 0.5em minus 0.4em\relax IOS Press,
  2018, pp. 61--70.

\bibitem[Jagadeesh et~al.(2005)Jagadeesh, Pingali, and
  Varma]{jagadeesh2005sentence}
J.~Jagadeesh, P.~Pingali, and V.~Varma, ``Sentence extraction based single
  document summarization,'' \emph{International Institute of Information
  Technology, Hyderabad, India}, vol.~5, 2005.

\bibitem[Hirao et~al.(2002)Hirao, Isozaki, Maeda, and
  Matsumoto]{hirao2002extracting}
T.~Hirao, H.~Isozaki, E.~Maeda, and Y.~Matsumoto, ``Extracting important
  sentences with support vector machines,'' in \emph{COLING 2002: The 19th
  International Conference on Computational Linguistics}, 2002.

\bibitem[Sugathadasa et~al.(2017)Sugathadasa, Ayesha, de~Silva, Perera,
  Jayawardana, Lakmal, and Perera]{sugathadasa2017synergistic}
K.~Sugathadasa, B.~Ayesha, N.~de~Silva, A.~S. Perera, V.~Jayawardana,
  D.~Lakmal, and M.~Perera, ``{Synergistic Union of Word2Vec and Lexicon for
  Domain Specific Semantic Similarity},'' \emph{{IEEE International Conference
  on Industrial and Information Systems (ICIIS)}}, pp. 1--6, 2017.

\bibitem[{Supreme Court}(1977)]{1977lee}
{Supreme Court}, ``{Lee v. United States},'' \emph{{US}}, vol. 432, no. No.
  76-5187, p.~23, 1977.

\end{thebibliography}

\end{document}